\documentclass[sigconf, screen]{acmart}

\usepackage{microtype}
\usepackage{graphicx}
\usepackage{caption}
\usepackage{algpseudocode}
\usepackage{comment}
\usepackage{color}
\usepackage{booktabs} 
\usepackage{url}
\usepackage{subfigure}
\usepackage{multicol}
\usepackage{xspace}
\usepackage{enumitem} 
\usepackage{amsmath}
\usepackage{array}

\usepackage{diagbox}
\usepackage{multirow}
\usepackage{hyperref}
\usepackage{verbatim}

\usepackage{array}
\newcolumntype{L}[1]{>{\raggedright\let\newline\\\arraybackslash\hspace{0pt}}m{#1}}
\newcolumntype{C}[1]{>{\centering\let\newline\\\arraybackslash\hspace{0pt}}m{#1}}
\newcolumntype{R}[1]{>{\raggedleft\let\newline\\\arraybackslash\hspace{0pt}}m{#1}}

\AtBeginDocument{%
  \providecommand\BibTeX{{%
    \normalfont B\kern-0.5em{\scshape i\kern-0.25em b}\kern-0.8em\TeX}}}

\copyrightyear{2020}
\acmYear{2020}
\setcopyright{acmcopyright}
\acmConference[ESEC/FSE '20]{Proceedings of the 28th ACM Joint European Software Engineering Conference and Symposium on the Foundations of Software Engineering}{November 8--13, 2020}{Virtual Event, USA}
\acmBooktitle{Proceedings of the 28th ACM Joint European Software Engineering Conference and Symposium on the Foundations of Software Engineering (ESEC/FSE '20), November 8--13, 2020, Virtual Event, USA}
\acmPrice{15.00}
\acmDOI{10.1145/3368089.3409691}
\acmISBN{978-1-4503-7043-1/20/11}

\begin{document}

\title{Object Detection for Graphical User Interface: Old Fashioned or Deep Learning or a Combination?}

\author{Jieshan Chen}
\email{Jieshan.Chen@anu.edu.au}
\affiliation{%
  \institution{Australian National University}
  \country{Australia}
  }
\author{Mulong Xie}
\email{u6462764@anu.edu.au}
\affiliation{%
  \institution{Australian National University}
  \country{Australia}
}

\author{Zhenchang Xing}
\email{Zhenchang.Xing@anu.edu.au}
\affiliation{%
  \institution{Australian National University}
  \country{Australia}
}
\additionalaffiliation{%
 \institution{Data61, CSIRO}
}

\author{Chunyang Chen}
\email{Chunyang.Chen@monash.edu}
\affiliation{%
  \institution{Monash University}
  \country{Australia}
}
\authornote{Corresponding author.}

\author{Xiwei Xu}
\email{Xiwei.Xu@data61.csiro.au}
\affiliation{%
  \institution{Data61, CSIRO}
}

\author{Liming Zhu}
\email{Liming.Zhu@data61.csiro.au}
\affiliation{%
  \institution{Data61, CSIRO}
  \country{Australia}
}
\additionalaffiliation{%
    \institution{University of New South Wales}
}

\author{Guoqiang Li}
\email{Li.G@sjtu.edu.cn}
\affiliation{%
  \institution{Shanghai Jiao Tong University}
  \country{China}
}

\renewcommand{\shortauthors}{Chen, J., Xie, M., Xing, Z., Chen, C., Xu, X., Zhu, L. and Li, G.}

\begin{abstract}
 Detecting Graphical User Interface (GUI) elements in GUI images is a domain-specific object detection task.
 It supports many software engineering tasks, such as GUI animation and testing, GUI search and code generation.
 Existing studies for GUI element detection directly borrow the mature methods from computer vision (CV) domain, including old fashioned ones that rely on traditional image processing features (e.g., canny edge, contours), and deep learning models that learn to detect from large-scale GUI data.
 Unfortunately, these CV methods are not originally designed with the awareness of the unique characteristics of GUIs and GUI elements and the high localization accuracy of the GUI element detection task.
 We conduct the first large-scale empirical study of seven representative GUI element detection methods on over 50k GUI images to understand the capabilities, limitations and effective designs of these methods.
 This study not only sheds the light on the technical challenges to be addressed but also informs the design of new GUI element detection methods.
 We accordingly design a new GUI-specific old-fashioned method for non-text GUI element detection which adopts a novel top-down coarse-to-fine strategy, and incorporate it with the mature deep learning model for GUI text detection.
 Our evaluation on 25,000 GUI images shows that our method significantly advances the start-of-the-art performance in GUI element detection.

\end{abstract}


\maketitle

\section{Introduction}
\label{sec:introduction}

GUI allows users to interact with software applications through graphical elements such as widgets, images and text.
Recognizing GUI elements in a GUI is the foundation of many software engineering tasks, such as GUI automation and testing~\cite{yeh2009sikuli, qian2020roscript, white2019improving, bernal2020translating}, supporting advanced GUI interactions~\cite{dixon2010prefab, banovic2012waken}, GUI search~\cite{deka2017rico, reiss2018seeking}, and code generation~\cite{nguyen2015reverse, moran2018machine, chen2018ui}.
Recognizing GUI elements can be achieved by instrumentation-based or pixel-based methods.
Instrumentation-based methods~\cite{lin2016sensing,pongnumkul2011pause, bao2015activityspace} are intrusive and requires the support of accessibility APIs~\cite{windowsAccessibilityApi, androidAccessibilityAPI} or runtime infrastructures~\cite{windowsspy,AndroidGUIAnimator} that expose information about GUI elements within a GUI.
In contrast, pixel-based methods directly analyze the image of a GUI, and thus are non-intrusive and generic.
Due to the cross-platform characteristics of pixel-based methods, they can be widely used for novel applications such as robotic testing of touch-screen applications~\cite{qian2020roscript}, linting of GUI visual effects~\cite{zhao2020seenomaly} in both Android and IOS.

Pixel-based recognition of GUI elements in a GUI image can be regarded as a domain-specific object detection task.
Object detection is a computer-vision technology that detects instances of semantic objects of a certain class (such as human, building, or car) in digital images and videos.
It involves two sub-tasks: \textit{region detection or proposal} - locate the bounding box (bbox for short) (i.e., the smallest rectangle region) that contains an object, and \textit{region classification} - determine the class of the object in the bounding box.
Existing object-detection techniques adopt a bottom-up strategy: starts with primitive shapes and regions (e.g., edges or contours) and aggregate them progressively into objects.
Old-fashioned techniques~\cite{nguyen2015reverse, moran2018machine, swearngin2018rewire} relies on image features and aggregation heuristics generated by expert knowledge, while deep learning techniques~\cite{ren2015faster, redmon2018yolov3, duan2019centernet} use neural networks to learn to extract features and their aggregation rules from large image data.

\begin{figure}
	\centering \includegraphics[width=0.45\textwidth]{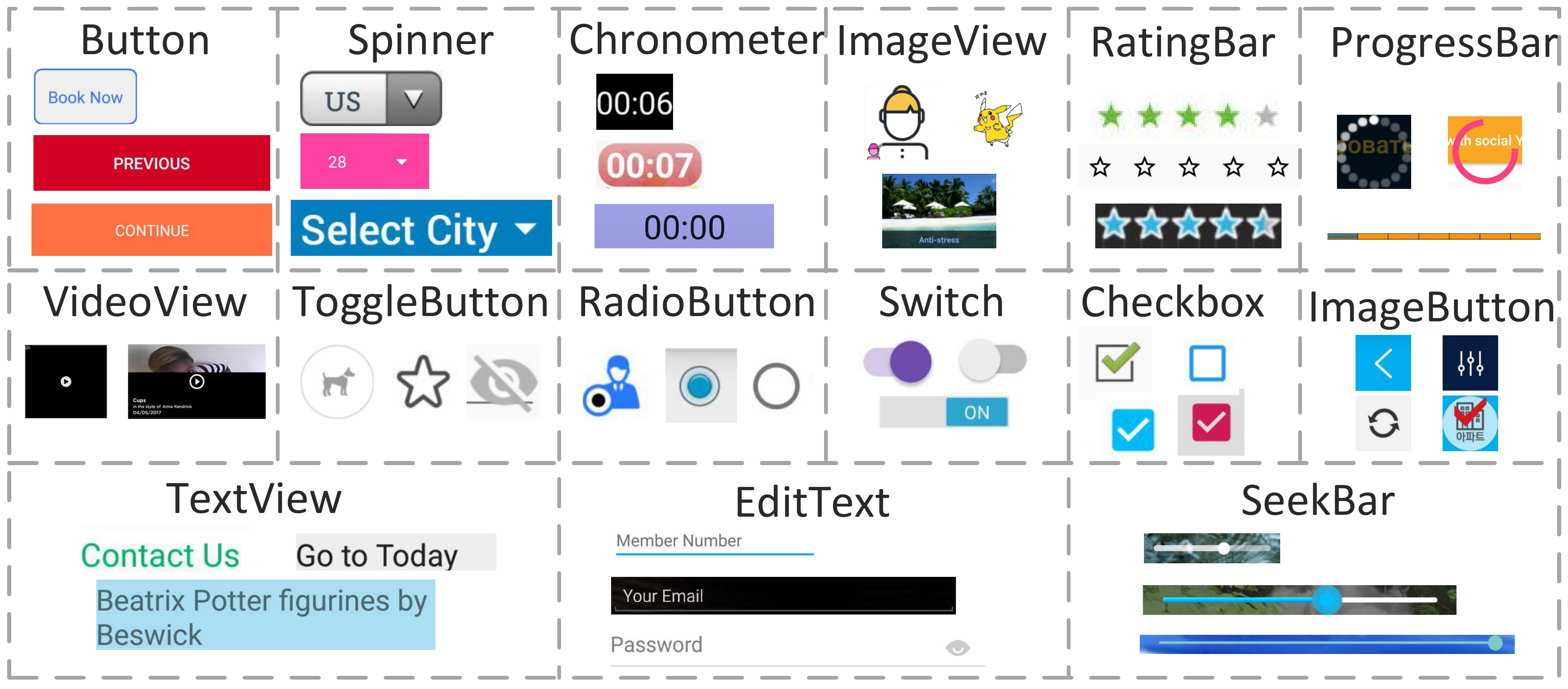}
	\vspace{-4.2mm}
	\caption{Characteristics of GUI elements: large in-class variance and high cross-class similarity }
	\label{fig:element_examples}
	\vspace{-9.8mm}
\end{figure}

GUI elements can be broadly divided into text elements and non-text elements (see Figure~\ref{fig:element_examples} for the examples of Android GUI elements).
Both old-fashioned techniques and deep learning models have been applied for GUI element detection~\cite{nguyen2015reverse, xianyu_blog, moran2018machine, white2019improving, chen2019gallery}.
As detailed in Section~\ref{sec:problemscope}, considering the image characteristics of GUIs and GUI elements, the high accuracy requirement of GUI-element region detection, and the design rationale of existing object detection methods, we raise a set of research questions regarding the effectiveness features and models originally designed for generic object detection on GUI elements, the region detection accuracy of statistical machine learning models, the impact of model architectures, hyperparameter settings and training data, and the appropriate ways of detecting text and non-text elements.

These research questions have not been systematically studied.
First, existing studies~\cite{nguyen2015reverse, moran2018machine, white2019improving} evaluate the accuracy of GUI element detection by only a very small number (dozens to hundreds) of GUIs.
The only large-scale evaluation is GUI component design gallery~\cite{chen2019gallery}, but it tests only the default anchor-box setting (i.e. a predefined set of bboxes) of Faster RCNN~\cite{ren2015faster} (a two-stage model).
Second, none of existing studies (including~\cite{chen2019gallery}) have investigated the impact of training data size and anchor-box setting on the performance of deep learning object detection models.
Furthermore, the latest development of anchor-free object detection has never been attempted.
Third, no studies have compared the performance of different methods, for example old fashioned versus deep learning, or different styles of deep learning (e.g., two stage versus one stage, anchor box or free).
Fourth, GUI text is simply treated by Optical Character Recognition (OCR) techniques, despite the significant difference between GUI text and document text that OCR is designed for.

To answer the raised research questions, we conduct the first large-scale, comprehensive empirical study of GUI element detection methods, involving a dataset of 50,524 GUI screenshots extracted from 8,018 Android mobile applications (see Section~\ref{sec:dataset}), and two representative old-fashioned methods (REMAUI~\cite{nguyen2015reverse} and Xianyu~\cite{xianyu_blog}) and three deep learning models (Faster RCNN~\cite{ren2015faster}, YOLOv3~\cite{redmon2018yolov3} and CenterNet~\cite{duan2019centernet}) that cover all major method styles (see Section~\ref{sec:baselines}).
Old-fashioned detection methods perform poorly (REMAUI F1=0.201 and Xianyu F1=0.154 at IoU$>$0.9) for non-text GUI element detection.
IoU is the intersection area over union area of the detected bounding box and the ground-truth box.
Deep learning methods perform much better than old-fashioned methods, and the two-stage anchor-box based Faster RCNN performs the best (F1=0.438 at IoU>0.9), and demands less training data.
However, even Faster RCNN cannot achieve a good balance of the coverage of the GUI elements and the accuracy of the detected bounding boxes.

It is surprising that anchor-box based models are robust to the anchor-box settings, and merging the detection results by different anchor-box settings can improve the final performance.
Our study shows that detecting text and non-text GUI elements by a single model performs much worse than by a dedicated text and non-text model respectively.
GUI text should be treated as scene text rather than document text, and the state-of-the-art deep learning scene text model EAST~\cite{zhou2017east} (pretrained without fine tuning) can accurately detect GUI text.

Inspired by these findings, we design a novel approach for GUI element detection.
For non-text GUI element detection, we adopt the simple two-stage architecture: perform region detection and region classification in a pipeline.
For non-text region detection, we prefer the simplicity and the bounding-box accuracy of old-fashioned methods.
By taking into account the unique boundary, shape, texture and layout characteristics of GUI elements, we design a novel old-fashioned method with a top-down coarse-to-fine detection strategy, rather than the current bottom-up edge/contour aggregation strategy in existing methods~\cite{nguyen2015reverse, xianyu_blog}.
For non-text region classification and GUI text detection, we adopt the mature, easy-to-deploy ResNet50 image classifier~\cite{he2016deep} and the EAST scene text detector~\cite{zhou2017east}, respectively.
By a synergy of our novel old-fashioned methods and existing mature deep learning models, our new method achieves 0.573 in F1 for all GUI elements, 0.523 in F1 for non-text GUI elements, and 0.516 in F1 for text elements in a large-scale evaluation with 25,000 GUI images, which significantly outperform existing old-fashioned methods, and outperform the best deep learning model by 19.4\% increase in F1 for non-text elements and 47.7\% increase in F1 for all GUI elements.

This paper makes the following contributions:

\begin{itemize}[leftmargin=*]
	\item We perform the first systematic analysis of the problem scope and solution space of GUI element detection, and identify the key challenges to be addressed, the limitations of existing solutions, and a set of unanswered research questions.
	
	\item We conduct the first large-scale empirical study of seven representative GUI element detection methods, which systematically answers the unanswered questions. We identify the pros and cons of existing methods which informs the design of new methods for GUI element detection.
	
	\item We develop a novel approach that effectively incorporates the advantages of different methods and achieves the state-of-the-art performance in GUI element detection.
\end{itemize}

\begin{table*}
	\centering
	\caption{Existing solutions for non-text GUI element detection and their limitations}
	\vspace{-3.5mm}
	\begin{tabular}{| L{1.5cm} | L{2.3cm} | L{7.5cm} | L{5cm} | }
		\hline
		\textbf{Style} & \textbf{Method} &  \textbf{Region Detection}    & \textbf{Region Classification}     \\
		\hline
		\multirow{2}{*}{}{\textbf{Old Fashioned}} & Edge/contour aggregation~\cite{nguyen2015reverse, xianyu_blog, moran2018machine}        &
		$\bullet$ Detect primitive edges and/or regions, and merge them into larger regions (windows or objects) \newline
		$\bullet$ Merge with text regions recognized by OCR \newline
		$\bullet$ Ineffective for artificial GUI elements (e.g., images)& Heuristically distinguish image, text, list, container~\cite{nguyen2015reverse, xianyu_blog}. Can be enhanced by a CNN classification like in ~\cite{moran2018machine}\\
		\cline{2-4}
		& Template matching~\cite{bao2015scvripper, qian2020roscript, yeh2009sikuli, dixon2010prefab}  & \multicolumn{2}{L{12.5cm}|}{$\bullet$ Depend on manual feature engineering (either sample images or abstract prototypes) \newline
			$\bullet$ Match samples/prototypes to detect object bounding box and class at the same time   \newline
			$\bullet$ Only applicable to simple and standard GUI elements (e.g., button, checkbox)  \newline
			$\bullet$ Hard to apply to GUI elements with large variance of visual features}     \\
		\hline
		\multirow{3}{*}{}{\textbf{Deep Learning}} & Anchor-box, two stage~\cite{ren2015faster, chen2019gallery} & $\bullet$ Must define anchor boxes  \newline
		$\bullet$ Pipeline region detection and region classification  \newline
		$\bullet$ Gallery D.C.~\cite{chen2019gallery} is the only work that tests the Faster RCNN on large-scale real GUIs, but it uses default settings & A CNN classifier for region classification, trained jointly with region proposal network  \\
		\cline{2-4}
		& Anchor-box, one stage~\cite{redmon2018yolov3, white2019improving}   & \multicolumn{2}{p{12.5cm}|}{ $\bullet$ YOLOv2~\cite{redmon2017yolo9000} and YOLOv3~\cite{redmon2018yolov3} uses k-means to determine anchor boxes (k is user-defined) \newline
			$\bullet$ Simultaneously region detection and region classification   \newline
			$\bullet$ \cite{white2019improving} uses YOLOv2; trains and tests on artificial desktop GUIs; only tests on 250 real GUIs}  \\
		\cline{2-4}
		& Anchor free~\cite{duan2019centernet} & \multicolumn{2}{c|}{Never applied}  \\
		\hline
	\end{tabular}
\vspace{-4.3mm}
	\label{tab:solution_space_nontext}
\end{table*}

\section{Problem Scope and Solution Space}
In this section, we identify the unique characteristics of GUIs and GUI elements, which have been largely overlooked when designing or choosing GUI element detection methods (Section~\ref{sec:problemscope}).
We also summarize representative methods for GUI element detection and point out the challenges that the unique characteristics of GUIs and GUI elements pose to these methods (Section~\ref{sec:solutionspace}).

\subsection{Problem Scope}
\label{sec:problemscope}

Figure~\ref{fig:element_examples} and Figure~\ref{fig:comparison} shows examples of GUI elements and GUIs in our dataset.
We observe two element-level characteristics: large in-class variance and high cross-class similarity, and two GUI-level characteristics: packed scene and close-by elements, and mix of heterogeneous objects.
In face of these characteristics, GUI element detection must achieve high accuracy on region detection.

\textbf{Large in-class variance:}
GUI elements are artificially designed, and their properties (e.g., height, width, aspect ratio and textures) depend on the content to display, the interaction to support and the overall GUI designs.
For example, the width of Button or EditText depends on the length of displayed texts.
ProgressBar may have different styles (vertical, horizontal or circle).
ImageView can display images with any objects or contents.
Furthermore, different designers may use different texts, colors, backgrounds and look-and-feel, even for the same GUI functionality.
In contrast, physical-world objects, such as human, car or building, share many shape, appearance and physical constraints in common within one class.
Large in-class variance of GUI elements pose main challenge of accurate region detection of GUI elements.

\textbf{High cross-class similarity:}
GUI elements of different classes often have similar size, shape and visual features.
For example, Button, Spinner and Chronometer all have rectangle shape with some text in the middle.
Both SeekBar and horizontal ProgressBar show a bar with two different portions.
The visual differences to distinguish different classes of GUI elements can be subtle.
For example, the difference between Button and Spinner lies in a small triangle at the right side of Spinner, while a thin underline distinguishes EditText from TextView.
Small widgets are differentiated by small visual cues.
Existing object detection tasks usually deal with physical objects with distinct features across classes, for example, horses, trucks, persons and birds in the popular COCO2015 dataset~\cite{lin2014microsoft}.
High cross-class similarity affects not only region classification but also region detection by deep learning models, as these two subtasks are jointly trained.

\textbf{Mix of heterogeneous objects:}
GUIs display widgets, images and texts.
Widgets are artificially rendered objects.
As discussed above, they have large in-class variance and high cross-class similarity.
ImageView has simple rectangle shape but can display any contents and objects.
For the GUI element detection task, we want to detect the ImageViews themselves, but not any objects in the images.
However, the use of visual features designed for physical objects (e.g., canny edge~\cite{canny}, contour map~\cite{contour}) contradicts this goal.
In Figure~\ref{fig:comparison} and Figure~\ref{fig:ocr_examples}, we can observe a key difference between GUI texts and general document texts.
That is, GUI texts are often highly cluttered with the background and close to other GUI elements, which pose main challenge of accurate text detection.
These heterogeneous properties of GUI elements must be taken into account when designing GUI element detection methods.

\textbf{Packed scene and close-by elements:}
As seen in Figure~\ref{fig:comparison}, GUIs, especially those of mobile applications, are often packed with many GUI elements, covering almost all the screen space.
In our dataset (see Section~\ref{sec:dataset}), 77\% of GUIs contain more than seven GUI elements.
Furthermore, GUI elements are often placed close side by side and separated by only small padding in between.
In contrast, there are only an average of seven objects placed sparsely in an image in the popular COCO(2015) object detection challenge~\cite{lin2014microsoft}.
GUI images can be regarded as packed scenes.
Detecting objects in packed scenes is still a challenging task, because close-by objects interfere the accurate detection of each object's bounding box.

\textbf{High accuracy of region detection}
For generic object detection, a typical correct detection is defined loosely, e.g., by an IoU$>$ 0.5 between the detected bounding box and its ground truth (e.g., the PASCAL VOC Challenge standard~\cite{everingham2010pascal}), since people can recognize an object easily from major part of it.
In contrast, GUI element detection has a much stricter requirement on the accuracy of region detection.
Inaccurate region detection may not only result in inaccurate region classification, but more importantly it also significantly affects the downstream applications, for example, resulting in incorrect layout of generated GUI code, or clicking on the background in vain during GUI testing.
However, the above GUI characteristics make the accurate region detection a challenging task.
Note that accurate region classification is also important, but the difficulty level of region classification relies largely on the downstream applications.
It can be as simple as predicting if a region is tapable or editable for GUI testing, or if a region is a widget, image or text in order to wireframe a GUI, or which of dozens of GUI framework component(s) can be used to implement the region.

\subsection{Solution Space}
\label{sec:solutionspace}
We summarize representative methods for GUI element detection, and raise questions that have not been systematically answered.

\subsubsection{Non-Text Element Detection}
Table~\ref{tab:solution_space_nontext} summarizes existing methods for non-text GUI element detection.
By contrasting these methods and the GUI characteristics in Section~\ref{sec:problemscope}, we raise a series questions for designing effective GUI element detection methods.
We focus our discussion on region detection, which aims to distinguish GUI element regions from the background.
Region classification can be well supported by a CNN-based image classifier~\cite{moran2018machine}.

\textbf{The effectiveness of physical-world visual features.}
Old-fashioned methods for non-text GUI element detection rely on either edge/contour aggregation~\cite{nguyen2015reverse, xianyu_blog, moran2018machine} or template matching~\cite{bao2015scvripper, qian2020roscript, yeh2009sikuli, dixon2010prefab}.
Canny edge~\cite{canny} and contour map~\cite{contour} are primitive visual features of physical-world objects, which are designed to capture fine-grained texture details of objects.
However, they do not intuitively correspond to the shape and composition of GUI elements.
It is error-prone to aggregate these fine-grained regions into GUI elements, especially when GUIs contain images with physical-world objects.
Template matching methods improve over edge/contour aggregation by guiding the region detection and aggregation with high-quality sample images or abstract prototypes of GUI elements.
But this improvement comes with the high cost of manual feature engineering.
As such, it is only applicable to simple and standard GUI widgets (e.g., button and checkbox of desktop applications).
It is hard to apply template-matching method to GUI elements of mobile applications which have large variance of visual features.
Deep learning models~\cite{ren2015faster, chen2019gallery, redmon2018yolov3, white2019improving, duan2019centernet} remove the need of manual feature engineering by learning GUI element features and their composition from large numbers of GUIs.
\textit{How effective can deep learning models learn GUI element features and their composition in face of the unique characteristics of GUIs and GUI elements?}

\textbf{The accuracy of bounding box regression.}
Deep learning based object detection learns a statistical regression model to predict the bounding box of an object.
This regression model makes the prediction in the feature map of a high layer of the CNN, where one pixel stands for a pixel block in the original image.
\textit{Can such statistical regression satisfy the high-accuracy requirement of region detection, in face of large in-class variance of GUI element and packed or close-by GUI elements?}

\textbf{The impact of model architectures, hyperparameters and training data.}
Faster RCNN~\cite{ren2015faster} and YOLOv2~\cite{redmon2018yolov3}) have been applied to GUI element detection.
These two models rely on a set of pre-defined anchor boxes.
The number of anchor boxes and their height, width and aspect ratio are all the model hyperparameters, which are either determined heuristically~\cite{ren2015faster} or by clustering the training images using k-means and then using the metrics of the centroid images~\cite{redmon2018yolov3}
Considering large in-class variance of GUI elements, \textit{how sensitive are these anchor-box based models to the definition of anchor boxes, when they are applied to GUI element detection?}
Furthermore, the recently proposed anchor-free model (e.g., CenterNet~\cite{duan2019centernet}) removes the need of pre-defined anchor-boxes, but has never been applied to GUI element detection.
\textit{Can anchor-free model better deal with large in-class variance of GUI elements?}
Last but not least, the performance of deep learning models heavily depends on sufficient training data.
\textit{How well these models perform with different amount of training data?}

\subsubsection{Text Element Detection}
Existing methods either do not detect GUI texts or detect GUI texts separately from non-text GUI element detection.
They simply use off-the-shelf OCR tools (e.g., Tesseract~\cite{smith2007overview}) for GUI text detection.
OCR tools are designed for recognizing texts in document images, but GUI texts are very different from document texts.
\textit{Is OCR really appropriate for detecting GUI texts?}
\textit{Considering the cluttered background of GUI texts, would it better to consider GUI text as scene text? Can the deep learning scene text model effectively detect GUI texts?}
Finally, considering the heterogeneity of GUI widgets, images and texts, \textit{can a single model effectively detect text and non-text elements?}

\section{Empirical Study}
\label{sec:empirical_study}

To answer the above unanswered questions, we conduct the first large-scale empirical study of using both old-fashioned and deep learning methods for GUI element detection.
Our study is done on a dataset of 50,524 GUI screenshots from the Rico dataset~\cite{deka2017rico}, which were extracted from 8,018 Android mobile applications from 27 application categories.
Our study involves a systematic comparison of two old-fashioned methods, including the representative method REMAUI~\cite{nguyen2015reverse} in the literature and the method Xianyu~\cite{xianyu_blog} recently developed by the industry, and three popular deep learning methods that cover all major model design styles, including two anchor-box based methods - Faster RCNN~\cite{ren2015faster} (two stage style) and YOLO V3~\cite{redmon2018yolov3} (one stage style) and one one-stage anchor-free model CenterNet~\cite{duan2019centernet}.
For GUI text detection, we compare OCR tool Tesseract~\cite{smith2007overview} and scene text detector EAST~\cite{zhou2017east}, and compare separate and unified detection of text and non-text GUI elements.

\subsection{Research Questions}
As region classification can be well supported by a CNN-based image classifier~\cite{moran2018machine}, the study focuses on three research questions (RQs) on region detection in GUI element detection task:

\begin{itemize}[leftmargin=*]
    \item \textbf{RQ1 Performance}: How effective can different methods detect the region of non-text GUI elements, in terms of the accuracy of predicted bounding boxes and the coverage of GUI elements?

	\item \textbf{RQ2 Sensitivity}: How sensitive are deep learning techniques to anchor-box settings and amount of training data?

	\item \textbf{RQ3 Text detection}: Does scene text recognition fit better for GUI text detection than OCR technique? Which option, separated versus unified text and non-text detection, is more appropriate?

\end{itemize}

\subsection{Experiment Setup}

\subsubsection{Dataset}
\label{sec:dataset}

\begin{figure}
	\centering \includegraphics[width=0.4\textwidth]{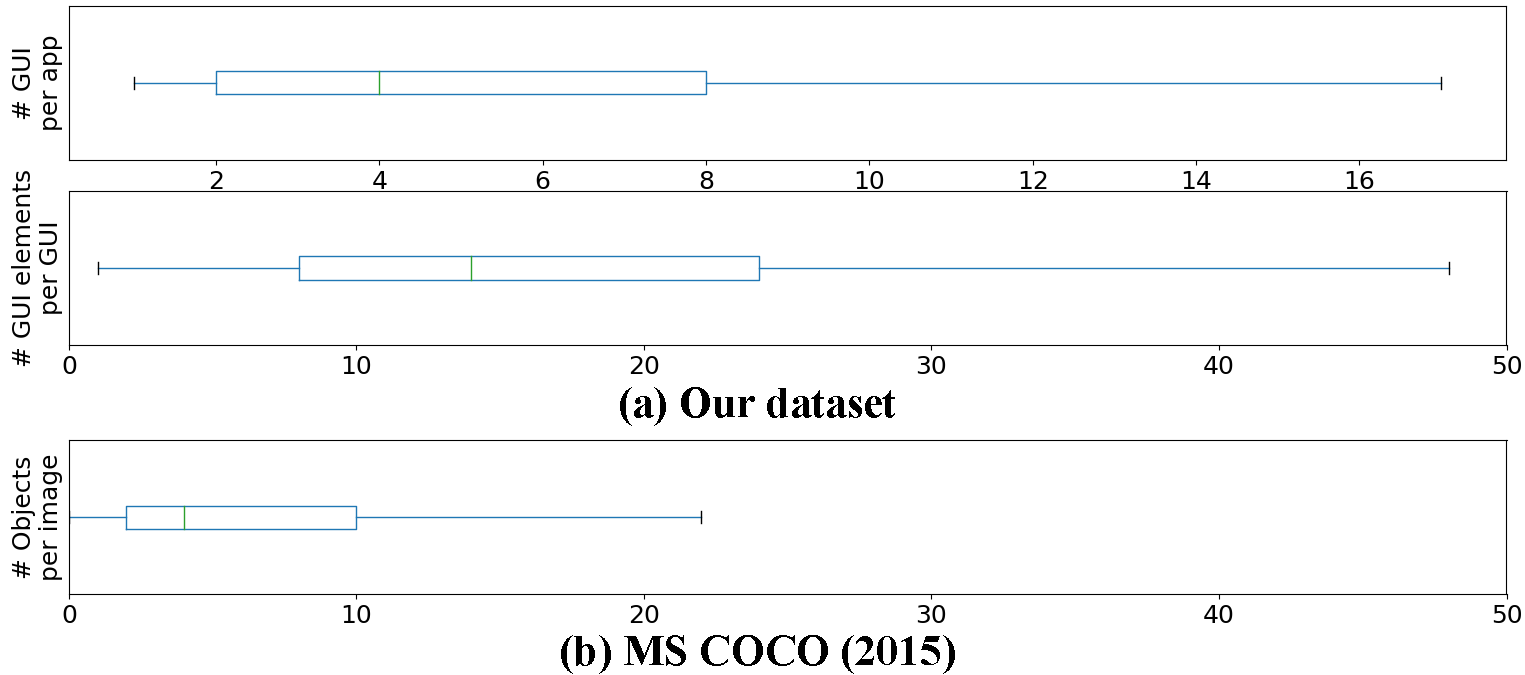}
	\vspace{-3.8mm}
	\caption{GUI elements distribution in our dataset}
	\label{fig:boxplots}
	\vspace{-7.8mm}
\end{figure}

We leverage Rico dataset~\cite{deka2017rico} to construct our experimental dataset.
In this study, we consider 15 types of commonly used GUI elements in the Android Platform (see examples in Figure~\ref{fig:element_examples}).
The Rico dataset contains 66,261 GUIs, and we filter out 15,737 GUIs.
Among them, 5,327 GUIs do not belong to the app itself, they are captured outside the app, such as Android home screen, a redirection to the social media login page (e.g. Facebook) or to a browser.
We identify them by checking whether the package name in the metadata for each GUI is different from the app's true package name.
2,066 GUIs do not have useful metadata, which only contain elements that describe the layout, or elements with invalid bounds, or do not have visible leaf elements.
709 of them do not contain any of the 15 elements.
The rest 7,635 GUIs are removed because they only contain text elements or non-text elements.
As a result, we obtain 50,524 GUI screenshots.
These GUIs are from 8,018 Android mobile applications of 27 categories.
These GUIs contain 923,404 GUI elements, among which 426,404 are non-text elements and 497,000 are text elements.
We remove the standard OS status and navigation bars from all GUI screenshots as they are not part of application GUIs.
We obtain the bounding-box and class of the GUI elements from the corresponding GUI metadata.
Figure~\ref{fig:boxplots} shows the distribution of GUIs per application and the number of GUI elements per GUI.
Compared with the number of objects per image in COCO2015, our GUI images are much more packed.
We split these 50,524 GUIs into train/validation/test dataset with a ratio of 8:1:1 (40K:5K:5k).
All GUIs of an application will be in only one split to avoid the bias of ``seen samples'' across training, validation and testing.
We perform 5-fold cross-validation in all the experiments.

\subsubsection{Baselines}\label{sec:baselines}
The baseline methods used in this study include:

\textbf{REMAUI}
\cite{nguyen2015reverse} 
detects text and non-text elements separately.
For text elements, it uses the OCR tool Tesseract~\cite{smith2007overview}.
For non-text elements, it detects the structural edge of GUI elements using Canny edge's detection~\cite{canny}, by using Gaussian filter to smooth the image and reduce noises, and then performing multi-stage filtering to identify the true edges in images.
After that, REMAUI performs edge merging, obtains the contours and obtains the bounding box of the GUI elements by merging partial overlapping regions.
We use the REMAUI tool~\cite{pix2app} provided by its authors in our experiments.

\textbf{Xianyu}
\cite{xianyu_blog} is a tool developed by the Alibaba to generate code from GUI images.
We only use the element detection part of this tool.
Xianyu binarizes the image and performs horizontal/vertical slicing, i.e., cutting the whole images horizontally/vertically in half, recursively to obtain the GUI elements.
It uses the edge detection method with Laplace filter, which highlights regions of rapid intensity change, to detect the edges and get contours in the binarized image.
Then, it leverages the flood fill algorithm~\cite{flood-fill}, which merge the qualified neighbor points gradually, to identify the connected regions and filters out the noise from the complex background.

\textbf{Faster RCNN}
\cite{ren2015faster} is a two-stage anchor-box-based deep learning technique for object detection.
It first generates a set of region proposals by a region proposal network (RPN), also called as region of interests (RoIs), which likely contain objects.
RPN uses a fixed set of user-defined boxes with different scales and aspect ratios (called anchor boxes) and computes these anchor boxes in each point in the feature map.
For each box, RPN then computes an objectness score to determine whether it contains an object or not, and regresses it to fit the actual bounding box of the contained object.
The second stage is a CNN-based image classifier that determines the object class in the RoIs.


\textbf{YOLOv3}
\cite{redmon2018yolov3} is an one-stage anchor-box-based object detection technique.
Different from the manually-defined anchor box of Faster-RCNN, YOLOv3 uses $k$-means method to cluster the ground truth bounding boxes in the training dataset, and takes the box scale and aspect ratio of the $k$ centroids as the anchor boxes.
It also extracts image features using CNN, and for each grid of the feature map, it generates a set of bounding boxes.
For each box, it computes the objectness scores, regresses the box coordinates and classifies the object in the bounding box at the same time.

\textbf{CenterNet}
\cite{duan2019centernet} is an one-stage anchor-free object detection technique.
Instead of generating bounding box based on the predefined anchor boxes, it predicts the position of the top-left and bottom-right corners and the center of an object, and then assembles them to get the bounding box of an object.
It computes the distance between each top-left corner and each bottom-right corner, and outputs the bounding box of the matched pair if the distance of them is smaller than a threshold and the center point of the bounding box has a certerness score higher than a threshold.


\textbf{Tesseract}
\cite{smith2007overview} is an OCR tool for document texts.
It consists of two steps: text line detection and text recognition.
Only the text line detection is relevant to our study.
The Tesseract's text line detection is old-fashioned.
It first converts the image into binary map, and then performs a connected component analysis to find the outlines of the elements.
These outlines are then grouped into blobs, which are further merged together.
Finally, it merges text lines that overlap at least half horizontally.

\textbf{EAST}
\cite{zhou2017east} is a deep learning technique to detect text in natural scenes.
An input image is first fed into a feature pyramid network.
EAST then computes six values for each point based on the final feature map, namely, an objectness score, top/left/bottom/right offsets and a rotation angle.
For this baseline, we directly use the pre-trained model to detect texts in GUIs without any fine-tuning.

\subsubsection{Model Training}
\label{sec:baseline_setting}
For faster RCNN, YOLOv3 and CenterNet, we initialize their parameters using the corresponding pre-trained models of COCO object detection dataset and fine-tune all parameters using our GUI training dataset.
We train each model for 160 iterations with a batch size of 8, and use Adam as the optimizer.
Faster RCNN uses ResNet-101~\cite{he2016deep} as the backbone.
YOLOv3 uses Darknet-53\cite{redmon2018yolov3} as the backbone.
CenterNet uses Hourglass-52~\cite{yang2017stacked} as the backbone.
For Xianyu and REMAUI, we perform the parameter tuning and use the best setting in all our evaluation.
We perform non-maximum suppression (NMS) to remove highly-duplicated predictions in all experiments.
It keeps the prediction with the highest objectness in the results and removes others if they have a IoU with the selected object over a certain value.
We find the best object confidence threshold for each model using the validation dataset.
All codes and models are released at our Github repository\footnote{\href{https://github.com/chenjshnn/Object-Detection-for-Graphical-User-Interface}{\textcolor{blue}{https://github.com/chenjshnn/Object-Detection-for-Graphical-User-Interface}}}.

\begin{figure}
	\centering
	\includegraphics[width=0.48\textwidth]{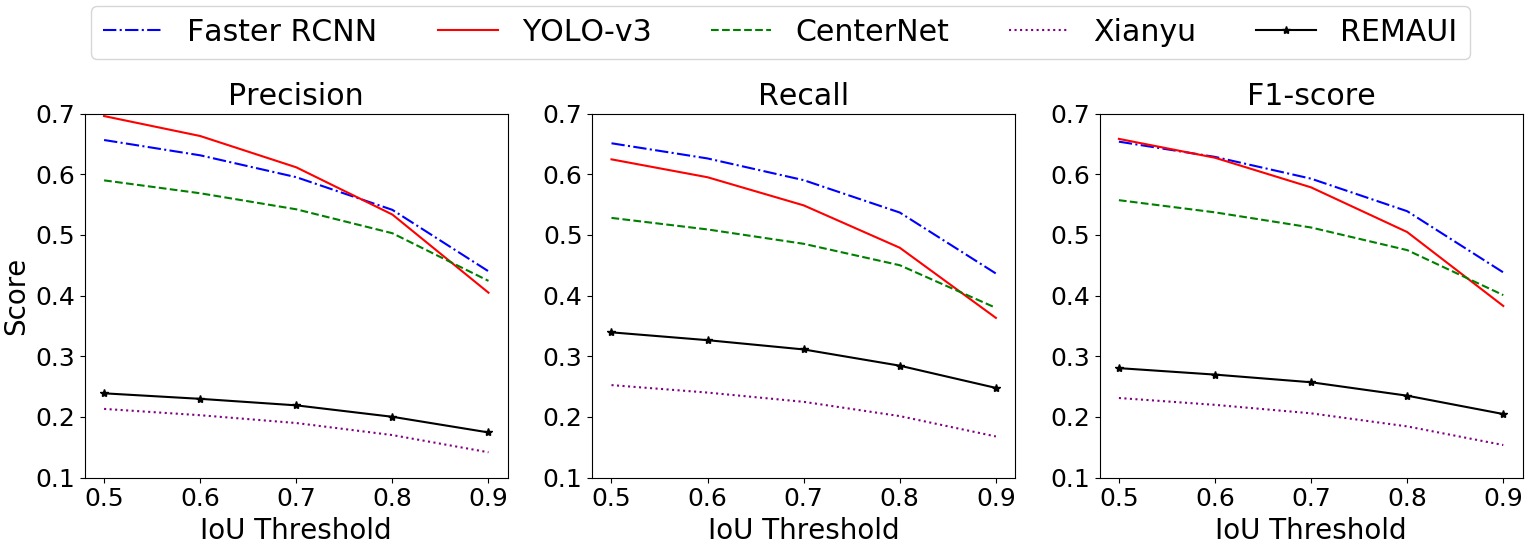}
	\vspace{-6mm}
	\caption{Performance at different IoU thresholds}
    \vspace{-5.5mm}
	\label{fig:different_ious}
\end{figure}

\subsubsection{Metrics}
For region detection evaluation, we ignore the class prediction results and only evaluate the ability of different methods to detect the bounding box of GUI elements.
We use precision, recall and F1-score to measure the performance of region detection.
Precision is $TP/(TP+FP) $ and recall is $TP/(TP+FN) $.
True positive (TP) refers to a detected bounding box which matches a ground truth box.
False positive (FP) refers to a detected box which does not match any ground truth boxes.
False negative (FN) refers to a ground truth bounding box which is not matched by any detected boxes.
We compute F1-score as: $ F1 = (2 \times Precision \times Recall)/(Precision + Recall) $.
TP is determined based on the Intersection over Union (IoU) of the two boxes.
IoU is calculated by dividing the intersection area $I$ of the two boxes $A$ and $B$ by the union area of the two boxes, i.e., $I/(A+B-I)$.
A detected box is considered as a TP if the highest IoU of this box with any ground-truth boxes in the input GUI image is higher than a predefined IoU threshold.
Each ground truth box can only be matched at most once and NMS technique is used to determine the optimal matching results.
Considering the high accuracy requirement of GUI element detection, we take the IoU threshold 0.9 in most of our experiments.

\subsection{Results - RQ1 Performance}
\label{sec:rq1_results}

This section reports the performance of five methods for detecting the regions of non-text GUI elements in a GUI.
In this RQ, Faster RCNN uses the customized anchor-box setting and YOLOv3 uses $k$=9 (see Section~\ref{sec:anchorboxsetting}).
The models are trained using all 40k training data and tested on 5k GUI images in 5-fold cross validation setting.

\subsubsection{Trade off between bounding-box accuracy and GUI-element coverage}
\label{sec:inaccuracy}
Figure~\ref{fig:different_ious} shows the performance of five methods at different IoU thresholds.
The F1-score of all deep learning models drop significantly when the IoU threshold increases from 0.5 to 0.9, with the 31\%, 45\% and 28\% decrease for Faster-RCNN, YOLOv3 and CenterNet respectively.
The bounding box of a RoI is predicted by statistical regression in the high-layer feature map of the CNN, where one pixel in this abstract feature map corresponds to a pixel block in the original image.
That is, a minor change of the predicted coordinates in the abstract feature map will lead to a large change in the exact position in the original image.
Therefore, deep learning models either detect more elements with loose bounding boxes or detect less elements with accurate bounding boxes.
In contrast, the F1-score of REMAUI and Xianyu does not drop as significantly as that of deep learning models as the IoU threshold increases, but their F1-scores are much lower than those of deep learning models.
This suggests that the detected element regions by these old-fashioned methods are mostly noise,
but when they do locate real elements, the detected bounding boxes are fairly accurate.

\subsubsection{Performance Comparison}\label{sec:performancecomparison}
We observe that if the detected bounding box has $<$0.9 IoU over the corresponding GUI element, not only does the box miss some portion of this element, but it also includes some portions of adjacent elements due to the packed characteristic of GUI design.
Therefore, we use IoU$>0.9$ as an acceptable accuracy of bounding box prediction.
Table~\ref{tab:region_proposal_detection} shows the overall performance of the five methods at IoU$>$0.9 threshold for detecting non-text GUI elements.

Xianyu performs the worst, with all metrics below 0.17.
We observe that Xianyu works fine for simple GUIs, containing some GUI elements on a clear or gradient background (e.g., Xianyu-(c)/(d) in Figure~\ref{fig:comparison}).
When the GUI elements are close-by or placed on a complex background image, Xianyu's slicing method and its background de-noising algorithms do not work well.
For example, in Xianyu-(a)/(b) in Figure~\ref{fig:comparison}, it misses most of GUI elements.
Xianyu performs slicing by the horizontal or vertical lines across the whole GUI.
Such lines often do not exist in GUIs, especially when they have complex background images (Xianyu-(a)) or the GUI elements are very close-by (Xianyu-(b)).
This results in many under-segmentation of GUI images and the misses of many GUI elements.
Furthermore, Xianyu sometimes may over-segment the background image (Xianyu-(a)), resulting in many noise non-GUI-element regions.

REMAUI performs better than Xianyu, but it is still much worse than deep learning models.
REMAUI suffers from similar problems as Xianyu, including ineffective background de-noising and over-segmentation.
It outperforms Xianyu because it merges close-by edges to construct bounding boxes instead of the simple slicing method by horizontal/vertical lines.
However, for GUIs with image background, its edge merging heuristics often fail due to the noisy edges of physical-world objects in the images.
As such, it often reports some non-GUI-element regions of the image as element regions, or erroneously merges close-by elements, as shown in Figure~\ref{fig:comparison}.
Furthermore, REMAUI merges text and non-text region heuristically, which are not very reliable either (see the text elements detected as non-text elements in REMAUI-(b)/(c)/(d)).

\begin{table}
	\centering
	\caption{Performance: non-text element detection (IoU$>$0.9)}
	\vspace{-3mm}

    \resizebox{0.4\textwidth}{12mm}{
	\begin{tabular}{l l c c c}
		\hline
		\textbf{Method} & \textbf{\#bbox} & \textbf{Precision} & \textbf{Recall} & \textbf{F1} \\
		\hline
		REMAUI      & 54,903  & 0.175 & 0.238 & 0.201 \\
		Xianyu      & 47,666  & 0.142 & 0.168 & 0.154 \\
		Faster-RCNN & 39,995  & \textbf{0.440} & \textbf{0.437} & \textbf{0.438} \\
		YOLOv3      & 36,191  & 0.405 & 0.363 & 0.383 \\
		CenterNet   & 36,096  & 0.424 & 0.380 & 0.401 \\
		\hline
	\end{tabular}
    }
	\vspace{-7mm}
	\label{tab:region_proposal_detection}
\end{table}

Deep learning models perform much better than old fashioned methods.
In Figure~\ref{fig:comparison}, we see that they all locate some GUI elements accurately, even those overlaying on the background picture.
These models are trained with large-scale data, ``see'' many sophisticated GUIs, and thus can locate GUI elements even in a noisy background.
However, we also observe that the detected bounding boxes by deep learning models may not be very accurate, as they are estimated by a statistical regression model.
The two-stage model Faster RCNN outperforms the other two one-stage models YOLOv3 and CenterNet.
As discussed in Section~\ref{sec:problemscope}, GUI elements have large in-class variance and high cross-class similarity.
Two stage models perform region detection and region classification in a pipeline so that the region detection and region classification are less mutually interfered, compared with one-stage models that perform region detection and region classification simultaneously.

\begin{table}
	\centering
	\caption{Impact of anchor-box settings (IoU$>$0.9)}
	\vspace{-3mm}
	\begin{tabular}{l c c c}
		\hline
		\textbf{Setting} & \textbf{Precision} & \textbf{Recall} & \textbf{F1} \\
		\hline
		Faster RCNN Default       & 0.433 & 0.410 & 0.421 \\
		Faster RCNN Customized    & 0.440 & 0.437 & 0.438 \\
		Faster RCNN Union         & 0.394 & 0.469 & 0.428 \\
		Faster RCNN Intersection  & \textbf{0.452} & \textbf{0.460} & \textbf{0.456} \\
		\hline
		YOLOv3 k=5         & 0.394 & 0.333 & 0.361 \\
		YOLOv3 k=9         & 0.405 & 0.363 & 0.383 \\
		YOLO Union         & 0.372 & 0.375 & 0.373 \\
		YOLO Intersection  & \textbf{0.430} & \textbf{0.424} & \textbf{0.427} \\
		\hline
	\end{tabular}
	\label{tab:anchors}
	\vspace{-4mm}
\end{table}

Between the two one-stage models, anchor-free CenterNet outperforms anchor-box-based YOLOv3 at IoU$>$0.9.
However, YOLOv3 performs better than CenterNet at lower IoU thresholds (see Figure~\ref{fig:different_ious}).
Anchor-free model is flexible to handle the large in-class variance of GUI elements and GUI texts (see more experiments on GUI text detection in Section~\ref{sec:rq4_results}).
However, as shown in Figure~\ref{fig:comparison}, this flexibility is a double-blade, which may lead to less accurate bounding boxes, or bound several elements in one box (e.g., CenterNet-(a)/(d)).
Because GUI elements are often close-by or packed in a GUI, CenterNet very likely assembles the top-left and bottom-right corners of different GUI elements together, which leads to the wrong bounding boxes.

\noindent
\fbox{
	\begin{minipage}{8.2cm} \emph{Deep learning models significantly outperform old-fashioned detection methods. Two-stage anchor-box-based models perform the best in non-text GUI element detection task. But it is challenging for the deep learning models to achieve a good balance between the accuracy of the detected bounding boxes and the detected GUI elements, especially for anchor-free models.}
\end{minipage}}

\begin{table}
    \centering
    \caption{Impact of amount of training data (IoU$>$0.9)}
    \vspace{-4mm}
    \begin{tabular}{l l c c c}
        \hline
        \textbf{Method} & \textbf{Size} & \textbf{Precision} & \textbf{Recall} & \textbf{F1} \\
        \hline
        \multirow{3}{*}{Faster-RCNN} & 2K  & 0.361 & 0.305 & 0.331 \\
                                     & 10K & 0.403 & 0.393 & 0.398 \\
                                     & 40K & \textbf{0.440} & \textbf{0.437} & \textbf{0.438} \\
        \hline
        \multirow{3}{*}{YOLOv3}      & 2K  & 0.303 & 0.235 & 0.265 \\
                                     & 10K & 0.337 & 0.293 & 0.313 \\
                                     & 40K & \textbf{0.405} & \textbf{0.363} & \textbf{0.383} \\
        \hline
        \multirow{3}{*}{CenterNet}   & 2K  & 0.319 & 0.313 & 0.316 \\
                                     & 10K & 0.328 & 0.329 & 0.329 \\
                                     & 40K & \textbf{0.424} & \textbf{0.380} & \textbf{0.401} \\
        \hline
    \end{tabular}
	\vspace{-5mm}
    \label{tab:dl_different_nums}
\end{table}

\subsection{Results - RQ2 Sensitivity}
\label{sec:rq2_results}
This section reports the sensitivity analysis of the deep learning models for region detection from two aspects: anchor-box settings and amount of training data.

\subsubsection{Anchor-Box Settings}
\label{sec:anchorboxsetting}

For Faster RCNN, we use two settings: the default setting (three anchor-box scales - 128, 256 and 512, and three aspect ratios - 1:1, 1:2 and 2:1); and the customized setting (five anchor-box scales - 32, 64, 128, 256 and 512, and four aspect (width:height) ratios - 1:1, 2:1, 4:1 and 8:1).
This customized setting is drawn from the frequent scales and aspect ratios of the GUI elements in our dataset.
Considering the size of GUI elements, we add two small scales 32 and 64.
Furthermore, we add two more aspect ratios to accommodate the large variance of GUI elements.
For YOLOv3, we use two $k$ settings: 5 and 9, which are commonly used in the literature.
YOLOv3 automatically derives anchor-box metrics from $k$ clusters of GUI images in the dataset.
All models are trained using 40k training data and tested on 5k GUI images.

Table~\ref{tab:anchors} shows the model performance (at IoU$>$0.9) of these different anchor-box settings.
It is somehow surprising that there is only a small increase in F1 when we use more anchor-box scales and aspect ratios.
We further compare the TPs of different anchor-box settings.
We find that 55\% of TPs overlap between the two settings for Faster RCNN, and 67\% of TPs overlap between the two settings for YOLOv3.
As the scales and aspect ratios of GUI elements follow standard distributions, using a smaller number of anchor boxes can still covers a large portion of the element distribution.

As different settings detect some different bounding boxes, we want to see if the differences may complement each other.
To that end, we adopt two strategies to merge the detected boxes by the two settings: union strategy and intersection strategy.
For two overlapped boxes, we take the maximum objectness of them, and then merge the two boxes by taking the union/intersection area for union/interaction strategy.
For the rest of the boxes, we directly keep them.
We find the best object confidence threshold for the combined results using the validation dataset.
The union strategy does not significantly affect the F1, which means that making the bounding boxes larger is not very useful.
In fact, for the boxes which are originally TPs by one setting, the enlarged box could even become FPs.
However, the intersection strategy can boost the performance of both Faster RCNN and  YOLOv3, achieving 0.456 and 0.427 in F1 respectively.
It is reasonable because the intersection area is confirmed by the two settings, and thus more accurate.

\subsubsection{Amount of Training Data.}\label{sec:trainingdatasize}
In this experiment, Faster RCNN uses the customized anchor-box setting and YOLOv3 uses $k$=9.
We train the models with 2K, 10K, 40K training data separately, and test the models on the same 5k GUI images.
Each 2k- or 10k experiment uses randomly selected 2k or 10k GUIs in the 40k training data.
As shown in Table~\ref{tab:dl_different_nums}, the performance of all models drops as the training data decreases.
This is reasonable because deep learning models cannot effectively learn the essential features of the GUI elements without sufficient training data.
The relative performance of the three models is consistent at the three training data sizes, with YOLOv3 always being the worst.
This indicates the difficulty in training one-stage anchor-box model.
Faster RCNN with 2k (or 10k) training data achieves the comparable or higher F1 than that of YOLOv3 and CenterNet with 10k (or 40k) training data.
This result further confirms that two-stage model fits better for GUI element detection tasks than one-stage model, and one-stage anchor-free model performs better than one-stage anchor-box model.

\noindent
\fbox{
	\begin{minipage}{8.2cm} \emph{Anchor-box settings do not significantly affect the performance of anchor-box-based models, because a small number of anchor boxes can cover the majority of GUI elements. Two-stage anchor-box-based model is the easiest to train, which requires one magnitude less training data to achieve comparable performance as one-stage model. One stage anchor-box model is the most difficult to train.}
\end{minipage}}

\subsection{Results - RQ3 Text Detection}
\label{sec:rq4_results}

\begin{table}
	\centering
	\caption{Text detection: separated versus unified processing}
	\vspace{-3mm}
    \resizebox{0.45\textwidth}{21mm}{
	\begin{tabular}{l | l l c c c}
		\hline
		\textbf{Method} & \multicolumn{2}{c}{\textbf{Element}} & \textbf{Precision} & \textbf{Recall} & \textbf{F1} \\
		\hline
		\multirow{4}{*}{Faster-RCNN} & \multicolumn{2}{c}{nontext-only} & 0.440 & 0.437 & 0.438  \\
		\cline{2-6}
		& \multirow{3}{*}{mix} & nontext   & \textbf{0.379} & \textbf{0.436} & \textbf{0.405}  \\
		&                      & text      & 0.275 & 0.250 & 0.262  \\
		&                      & both      & 0.351 & 0.359 & 0.355  \\
		\hline
		\multirow{4}{*}{YOLOv3}      & \multicolumn{2}{c}{nontext-only} & 0.405 & 0.363 & 0.383 \\
		\cline{2-6}
		& \multirow{3}{*}{mix} & non-text  & 0.325 & 0.347 & 0.335 \\
		&                      & text      & 0.319 & 0.263 & 0.288 \\
		&                      & both      & 0.355 & 0.332 & 0.343 \\
		\hline
		\multirow{4}{*}{CenterNet}   & \multicolumn{2}{c}{nontext-only} & 0.424 & 0.380 & 0.401 \\
		\cline{2-6}
		& \multirow{3}{*}{mix} & non-text  & 0.321 & 0.397 & 0.355 \\
		&                      & text      & \textbf{0.416} & \textbf{0.319} & \textbf{0.361} \\
		&                      & both      & \textbf{0.391} & \textbf{0.385} & \textbf{0.388} \\
		\hline
	\end{tabular}
}
	\vspace{-4.5mm}
	\label{tab:unified_models}
\end{table}

\subsubsection{Separated versus Unified Text/Non-Text Element Detection}
All existing works detect GUI text separately from non-text elements.
This is intuitive in that GUI text and non-text elements have very different visual features.
However, we were wondering if this is a must or text and non-text elements can be reliably detected by a single model.
To answer this, we train Faster RCNN, YOLOv3 and CenterNet to detect both text and non-text GUI elements.
Faster RCNN uses the customized anchor-box setting and YOLOv3 uses $k$=9.
The model is trained with 40k data and tested on 5k GUI images.
In this RQ, both non-text and text elements in GUIs are used for model training and testing.

Table~\ref{tab:unified_models} shows the results.
When trained to detect text and non-text elements together, Faster RCNN still performs the best in terms of detecting non-text elements.
But the performance of all three models for detecting non-text elements degrades, compared with the models trained to detect non-text elements only.
This indicates that mixing the learning of text and non-text element detection together interfere with the learning of detecting non-text elements.
CenterNet performs much better for detecting text elements than Faster RCNN and YOLOv3, which results in the best overall performance for the mixed text and non-text detection.
CenterNet is anchor-free, which makes it flexible to handle large variance of text patterns.
So it has comparable performance for text and non-text elements.
In contrast, anchor-box-based Faster RCNN and YOLOv3 are too rigid to reliably detect text elements.
However, the performance of CenterNet in detecting text elements is still poor.
Text elements always have space between words and lines.
Due to the presence of these spaces, CenterNet often detects a partial text element or erroneously groups separate text elements as one element when assembling object corners.

\subsubsection{OCR versus Scene Test Recognition}
Since it is not feasible to detect text and non-text GUI elements within a single model, we want to investigate what is the most appropriate method for GUI text detection.
All existing works (e.g., REMAUI, Xianyu) simply use OCR tool like Tesseract.
We observe that GUI text is more similar to scene text than to document text.
Therefore, we adopt a deep learning scene text recognition model EAST for GUI text detection, and compare it with Tesseract.
We directly use the pre-trained EAST model without any fine tuning on GUI text.

As shown in Table~\ref{tab:ocr_results}, EAST achieves 0.402 in precision, 0.720 in recall and 0.516 in F1, which is significantly higher than Tesseract (0.291 in precision, 0.518 in recall and 0.372 in F1).
Both Xianyu and REMAUI perform some post-processing of the Tesseract's OCR results in order to filter out false positives.
But it does not significantly change the performance of GUI text detection.
As EAST is specifically designed for scene text recognition, its performance is significantly better than using generic object detection models for GUI text detection (see Table~\ref{tab:unified_models}).
EAST detects almost all texts in a GUI, including those on the GUI widgets (e.g., the button labels in Figure~\ref{fig:ocr_examples}(c)).
However, those texts on GUI widgets are considered as part of the widgets in our ground-truth data, rather than stand-alone texts.
This affects the precision of EAST against our ground-truth data, even though the detected texts are accurate.

Figure~\ref{fig:ocr_examples} presents some detection results.
Tesseract achieves the comparable results as EAST only for the left side of Figure~\ref{fig:ocr_examples}(d), where text is shown on a white background just like in a document.
From all other detection results, we can observe the clear advantages of treating GUI text as scene text than as document text.
First, EAST can accurately detect text in background image (Figure~\ref{fig:ocr_examples}(a)), while Tesseract outputs many inaccurate boxes in such images.
Second, EAST can detect text in a low contrast background (Figure~\ref{fig:ocr_examples}(b)), while Tesseract often misses such texts.
Third, EAST can ignore non-text elements (e.g., the bottom-right switch buttons in Figure~\ref{fig:ocr_examples}(b), and the icons on the left side of Figure~\ref{fig:ocr_examples}(d)), while Tesseract often erroneously detects such non-text elements as text elements.

\noindent
\fbox{
	\begin{minipage}{8.2cm} \emph{GUI text and non-text elements should be detected separately. Neither OCR techniques nor generic object detection models can reliably detect GUI texts. As GUI texts have the characteristics of scene text, the deep learning scene text recognition model can be used (even without fine-tuning) to accurately detect GUI texts.}
\end{minipage}}

\begin{table}
	\centering
	\caption{Text detection: OCR versus scene text}
	\vspace{-3mm}
    \resizebox{0.3\textwidth}{9mm}{
	\begin{tabular}{l c c c}
		\hline
		\textbf{Method} & \textbf{Precision} & \textbf{Recall} & \textbf{F1} \\
		\hline
		Tesseract       & 0.291 & 0.518 & 0.372 \\
		EAST            & \textbf{0.402} & \textbf{0.720} & \textbf{0.516} \\
		REMAUI          & 0.297 & 0.489 & 0.369 \\
		Xianyu          & 0.272 & 0.481 & 0.348 \\
		\hline
	\end{tabular}
}
	\vspace{-4mm}
	\label{tab:ocr_results}
\end{table}

\section{A Novel Approach}
Based on the findings in our empirical study, we design a novel approach for GUI element detection.
Our approach combines the simplicity of old-fashioned computer vision methods for non-text-element region detection, and the mature, easy-to-deploy deep learning models for region classification and GUI text detection (Section~\ref{sec:approachdesign}).
This synergy achieves the state-of-the-art performance for the GUI element detection task (Section~\ref{sec:evaluation}).

\subsection{Approach Design}
\label{sec:approachdesign}
Our approach detects non-text GUI elements and GUI texts separately.
For GUI text detection, we simply use the pre-trained state-of-the-art scene text detector EAST~\cite{zhou2017east}.
For non-text GUI element detection, we adopt the two-stage design, i.e, perform region detection and region classification in a pipeline.
For region detection, we develop a novel old-fashioned method with a top-down coarse-to-fine strategy and a set of GUI-specific image processing algorithms.
For region classification, we fine-tune the pre-trained ResNet50 image classifier~\cite{he2016deep} with GUI element images.

\subsubsection{Region Detection for Non-Text GUI Elements}
According to the performance and sensitivity experiments results, we do not want to use generic deep learning object detection models~\cite{redmon2017yolo9000, ren2015faster, duan2019centernet}.
First, they demand sufficient training data, and different model designs require different scale of training data to achieve stable performance.
Furthermore, the model performance is still less optimal even with a large set of training data, and varies across different model designs.
Second, the nature of statistical regression based region detection cannot satisfy the high accuracy requirement of GUI element detection.
Unlike generic object detection where a typical correct detection is defined loosely (e.g, IoU$>$0.5)~\cite{vocchallenge2010}),
detecting GUI elements is a fine-grained recognition task which requires a correct detection that covers the full region of the GUI elements as accurate as possible, but the region of non-GUI elements and other close-by GUI elements as little as possible.
Unfortunately, neither anchor-box based nor anchor-free models can achieve this objective, because they are either too strict or too flexible in face of large in-class variance of element sizes and texture, high cross-class shape similarity, and the presence of close-by GUI elements.

\begin{figure}[t]
	\centering
	\includegraphics[width=0.4\textwidth]{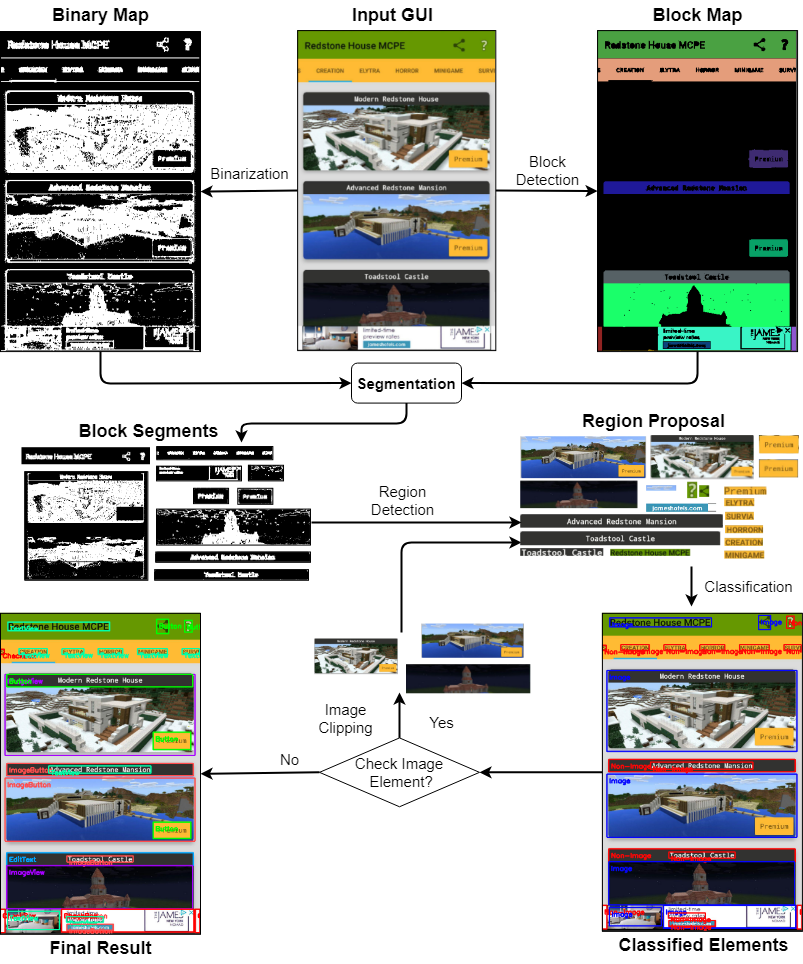}
	\vspace{-3.5mm}
	\caption{Our method for non-text GUI element detection }
	\vspace{-7mm}
	\label{fig:app_overview}
\end{figure}

Unlike deep learning models, old-fashioned methods~\cite{nguyen2015reverse, xianyu_blog} do not require any training which makes them easy to deploy.
Furthermore, when old fashioned methods locate some GUI elements, the detected bounding boxes are usually accurate, which is desirable.
Therefore, we adopt old-fashioned methods for non-text GUI-element region detection.
However, existing old-fashioned methods use a bottom-up strategy which aggregates the fine details of the objects (e.g., edge or contour) into objects.
This bottom-up strategy performs poorly, especially affected by the complex background or objects in the GUIs and GUI elements.
As shown in Figure \ref{fig:app_overview}, our method adopts a completely different strategy: top-down coarse-to-fine.
This design carefully considers the regularity of GUI layouts and GUI-element shapes and boundaries, as well as the significant differences between the shapes and boundaries of artificial GUI elements and those of physical-world objects.

\begin{figure}
	\centering
	\includegraphics[width=0.43\textwidth]{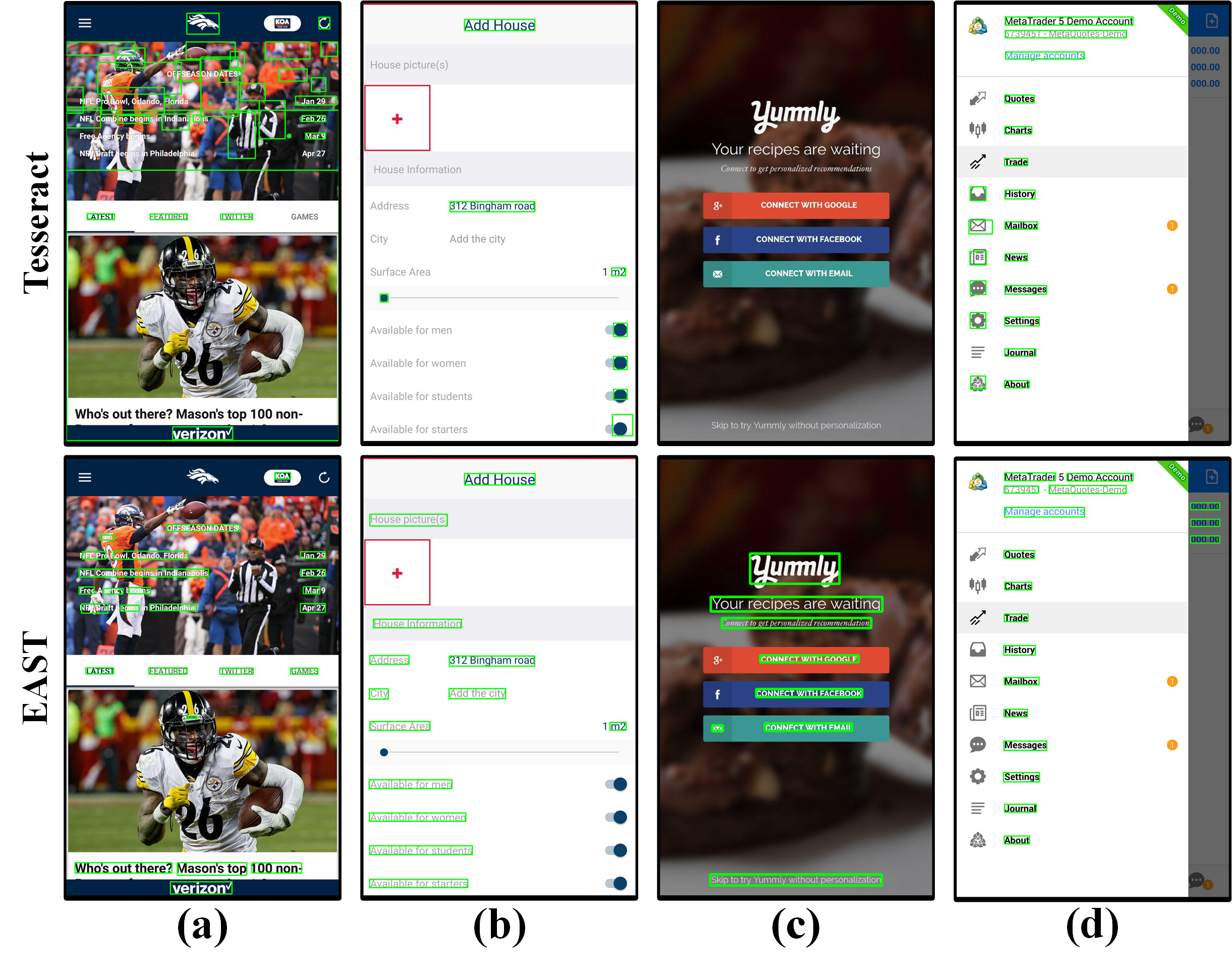}
	\vspace{-5.5mm}
	\caption{Examples: OCR versus scene text}
	\label{fig:ocr_examples}
	\vspace{-7.5mm}
\end{figure}

\begin{figure*}
	\centering
	\includegraphics[width=0.94\textwidth]{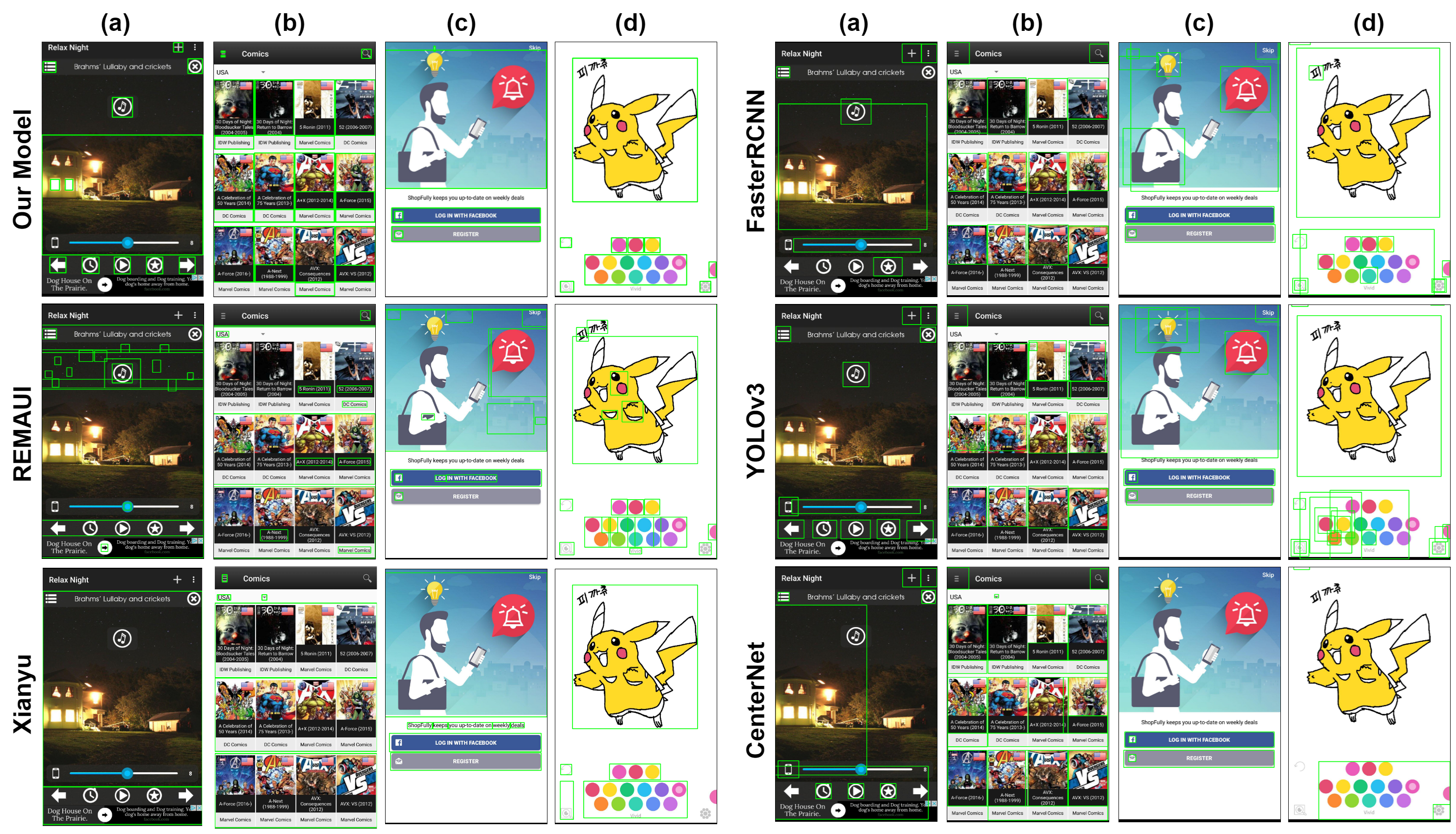}
	\vspace{-4.5mm}
	\caption{Region detection results for non-text GUI element: our method versus five baselines}
	\label{fig:comparison}
	\vspace{-5mm}
\end{figure*}

Our region detection method first detects the layout blocks of a GUI.
The intuition is that GUIs organize GUI elements into distinct blocks, and these blocks generally have rectangle shape.
Xianyu also detects blocks, but it assumes the presence of clear horizontal and vertical lines.
Our method does not make this naive assumption.
Instead, it first uses the flood-filling algorithm~\cite{flood-fill} over the grey-scale map of the input GUI to obtain the maximum regions with similar colors, and then uses the shape recognition~\cite{dp_approPoly} to determine if a region is a rectangle.
Each rectangle region is considered as a block.
Finally, it uses the
Suzuki's Contour tracing algorithm \cite{contour} to compute the boundary of the block and produce a block map.
In Figure \ref{fig:app_overview}, we show the detected block in different colors for the presentation clarity.
Note that blocks usually contain some GUI elements, but some blocks may correspond to a particular GUI element.

Next, our method generates a binary map of the input GUI, and for each detected block, it segments the corresponding region of the binary map.
Binarization simplifies the input image into a black-white image, on which the foreground GUI elements can be separated from the background.
Existing methods~\cite{nguyen2015reverse, xianyu_blog} perform binarization through Canny edge detection~\cite{canny} and Sobel edge detection~\cite{sobel_detection}, which are designed to keep fine texture details in nature scene images.
Unfortunately, this detail-keeping capability contradicts the goal of GUI element detection, which is to detect the shape of GUI elements, rather than their content and texture details.
For example, we want to detect an ImageView element no matter what objects are shown in the image (see Figure~\ref{fig:comparison}).
We develop a simple but effective binarization method based on the gradient map~\cite{gradient_map} of the GUI images.
A gradient map captures the change of gradient magnitude between neighboring pixels.
If a pixel has small gradient with neighboring pixels, it becomes black on the binary map, otherwise white.
As shown in Figure \ref{fig:app_overview}, the GUI elements stand out from the background in the binary map, either as white region on the black background or black region with white edge.

Our method uses the connected component labeling~\cite{ccl} to identify GUI element regions in each binary block segment.
It takes as input the binarized image and performs two-pass scanning to label the connected pixels.
As GUI elements can be any shape, it identifies a smallest rectangle box that covers the detected regions as the bounding boxes.
Although our binarization method does not keep many texture details of non-GUI objects, the shape of non-GUI objects (e.g., those buildings in the pictures) may still be present in the binary map.
These noisy shapes interfere existing bottom-up aggregation methods~\cite{canny, contour} for GUI element detection, which results in over-segmentation of GUI elements.
In contrast, our top-down detection strategy minimizes the influence of these non-GUI objects, because it uses relaxed grey-scale map to detect large blocks and then uses strict binary map to detect GUI elements.
If a block is classified as an image, our method will not further detect GUI elements in this block.

\begin{table}
	\centering
	\caption{Detection performance of our approach (IoU$>$0.9)}
	\vspace{-3.8mm}
    \resizebox{0.3\textwidth}{8mm}{
	\begin{tabular}{l c c c}
		\hline
		\textbf{Elements} & \textbf{Precision} & \textbf{Recall} & \textbf{F1} \\
		\hline
		non-text & 0.503 & 0.545 & 0.523 \\
		text & 0.589 & 0.547 & 0.516 \\
		both    & 0.539 & 0.612 & 0.573 \\
		\hline
	\end{tabular}
}
	\label{tab:ourapproachevaluation}
	\vspace{-4.5mm}
\end{table}

\begin{table}
	\centering
	\caption{Region classification results for TP regions}
	\vspace{-3.5mm}
    \resizebox{0.4\textwidth}{12mm}{
	\begin{tabular}{l| c c | c c}
		\hline
		& \multicolumn{2}{c|}{\textit{Non-text elements}}       & \multicolumn{2}{c}{\textit{All elements}} \\
		\hline
		\textbf{Method} & \textbf{\#bbox}  & \textbf{Accuracy}  & \textbf{\#bbox}  & \textbf{Accuracy} \\
		\hline
		FasterRCNN    & 18,577          & 0.68           &  34,915          & 0.68           \\
		YOLOv3        & 15,428          & 0.64           &  32,225          & 0.65           \\
		Centernet     & 16,072          & 0.68           &  36,803          & 0.66           \\
		Our method    & 21,977 & \textbf{0.86}  &  53,027 & \textbf{0.91}  \\
		\hline
	\end{tabular}
    }
	\vspace{-4mm}
	\label{tab:cnn_results}
\end{table}

\begin{table}
	\centering
	\caption{Overall results of object detection  (IoU $>$ 0.9)
	}
	\vspace{-4mm}
	\resizebox{0.48\textwidth}{15mm}{
		\begin{tabular}{l | c c c | c c c}
			\hline
			& \multicolumn{3}{c|}{\textit{Non-text elements}}       & \multicolumn{3}{c}{\textit{All elements}} \\
			\hline
			\textbf{Method} & \textbf{Precision} & \textbf{Recall} & \textbf{F1} & \textbf{Precision} & \textbf{Recall} & \textbf{F1} \\
			\hline
			Faster-RCNN     & 0.316 & 0.313 & 0.315 & 0.269 & 0.274 & 0.271 \\
			YOLOv3          & 0.274 & 0.246 & 0.260 & 0.258 & 0.242 & 0.249 \\
			CenterNet       & 0.302 & 0.270 & 0.285 & 0.284 & 0.280 & 0.282  \\
            Xianyu          & 0.122 & 0.145 & 0.133 & 0.270 & 0.405 & 0.324 \\
            REMAUI          & 0.151 & 0.205 & 0.173 & 0.296 & 0.449 & 0.357 \\
			Our method      & \textbf{0.431} & \textbf{0.469} & \textbf{0.449} & \textbf{0.490} & \textbf{0.557} & \textbf{0.524} \\
			\hline
		\end{tabular}
	}
	\vspace{-7.5mm}
	\label{tab:overall_performance}
\end{table}

\subsubsection{Region Classification for Non-Text GUI Elements}
For each detected GUI element region in the input GUI, we use a ResNet50 image classifier to predict its element type.
In this work, we consider 15 element types as shown in Figure~\ref{fig:element_examples}.
The Resnet50 image classifier is pre-trained with the ImageNet data.
We fine-tune the pre-trained model with 90,000 GUI elements (6,000 per element type) randomly selected from the 40k GUIs in our training dataset.

\subsubsection{GUI Text Detection}
Section~\ref{sec:rq4_results} shows that GUI text should be treated as scene text and be processed separately from non-text elements.
Furthermore, scene text recognition model performs much better than generic object detection models.
Therefore, we use the state-of-the-art deep-learning scene text detector EAST~\cite{zhou2017east} to detect GUI text.
As shown in Figure~\ref{fig:ocr_examples}(c), EAST may detect texts that are part of non-image GUI widgets (e.g., the text on the buttons).
Therefore, if the detected GUI text is inside the region of a non-image GUI widgets, we discard this text.

\subsection{Evaluation}
\label{sec:evaluation}

We evaluate our approach on the same testing data used in our empirical study by 5-fold cross-validation.
Table~\ref{tab:ourapproachevaluation} shows the region-detection performance for non-text, text and both types of elements.
For non-text GUI elements, our approach performs better than the best baseline Faster RCNN (0.523 versus 0.438 in F1).
For text elements, our approach is overall the same as EAST.
It is better than EAST in precision, because our approach discards some detected texts that are a part of GUI widgets.
But this degrades the recall.
For text and non-text elements as a whole, our approach performs better than the best baseline CenterNet (0.573 versus 0.388 in F1).

Figure~\ref{fig:comparison} shows the examples of the detection results by our approach and the five baselines.
Compared with REMAUI and Xianyu, our method detects much more GUI elements and much less noisy non-GUI element regions, because of our robust top-down coarse-to-fine strategy and GUI-specific image processing (e.g., connected component labeling rather than canny edge and contour).
Our method also detects more GUI elements than the three deep learning models.
Furthermore, it outputs more accurate bounding boxes and less overlapping bounding boxes, because our method performs accurate pixel analysis rather than statistical regression in the high-layer of CNN.
Note that deep learning models may detect objects in images as GUI elements, because there are GUI elements of that size and with similar visual features.
In contrast, our method detects large blocks that are images and treats such images as whole.
As such, our method suffers less over-segmentation problem.

For failure analysis of our model, we conclude three main reasons when our model fails.
First, same look and feel UI regions may correspond to different types of widgets, such as text label versus text button without border.
This is similar to the widget tappability issue studied in~\cite{swearngin2019modeling}.
Second, the repetitive regions in a dense UI (e.g., Figure~\ref{fig:comparison}(b)) often have inconsistent detection results.
Third, it is sometimes hard to determine whether a text region is a text label or part of a widget containing text, for example, the spinner showing USA at the top of Figure~\ref{fig:comparison}(b).
Note that these challenges affect all methods.
We leave them as our future work.

Table~\ref{tab:cnn_results} shows the region classification results of our CNN classifier and the three deep learning baselines.
The results consider only true-positive bounding boxes, i.e., the classification performance given the accurate element regions.
As text elements are outputted by EAST directly, we show the results for non-text elements and all elements.
We can see that our method outputs more true-positive GUI element regions, and achieves higher classification accuracy (0.86 for non-text elements and 0.91 for all elements, and the other three deep models achieves about 0.68 accuracy).
Our classification accuracy is consistent with~\cite{moran2018machine}, which confirms that the effectiveness of a pipeline design for GUI element detection.

Table~\ref{tab:overall_performance} shows the overall object detection results, i.e., the true-positive bounding box with the correct region classification over all detected element regions.
Among the three baseline models, Faster RCNN performs the best for non-text elements (0.315 in F1), but CenterNet, due to this model flexibility to handle GUI texts, achieves the best performance for all elements (0.282 in F1).
Compared with these three baselines, our method achieves much better F1 for both non-text elements (0.449) and all elements (0.524), due to its strong capability in both region detection and region classification.

\section{Related Work}

GUI design, implementation and testing are important software engineering tasks, to name a few, GUI code generation~\cite{chen2018ui, bielik2018robust, moran2018machine}, GUI search~\cite{zheng2019faceoff, huang2019swire, chen2019gallery, chen2020wireframe, chen2020tag}, GUI design examination~\cite{moran2018automated, swearngin2019modeling, zhao2020seenomaly}, reverse-engineering GUI dataset~\cite{deka2017rico, chen2019storydroid}, GUI accessibility~\cite{chen2020unblind}, GUI testing~\cite{qian2020roscript,bernal2020translating, li2019humanoid, white2019improving, liu2020owl} and GUI security~\cite{chen2019gui, xi2019deepintent}.
Many of these tasks require the detection of GUI elements.
As an example, the RQ4 in~\cite{white2019improving} shows exploiting exact widget locations by instrumentation achieves significantly higher branch coverage than predicted locations in GUI testing, but widget detection (by YOLOv2) can interact with widgets not detected by instrumentation.
Our work focuses on the foundational technique to improve widget detection accuracy, which opens the door to keep the advantage of widget detection while achieving the benefits of instrumentation in downstream applications like GUI testing.

Table~\ref{tab:solution_space_nontext} and Section~\ref{sec:baselines} summarizes the old-fashioned methods (e.g., REMAUI~\cite{nguyen2015reverse} and Xianyu~\cite{xianyu_blog}) designed for GUI element detection, the generic object detection models (Faster RCNN, YOLOv3 and CenterNet) applied for non-text GUI element detection, and the old-fashioned OCR tool Tesseract~\cite{smith2007overview} and the state-of-the-art scene text detector EAST~\cite{zhou2017east} for GUI text detection.
Our empirical study shows that old-fashioned methods perform poorly for both text and non-text GUI element detection.
Generic object detection models perform better than old-fashioned ones, but they cannot satisfy the high accuracy requirement of GUI element detection.
Our method advances the state-of-the-art in GUI element detection by effectively assembling the effective designs of existing methods and a novel GUI-specific old-fashioned region detection method.

Besides, some researchers apply image captioning to generate GUI code from GUI design~\cite{chen2018ui, beltramelli2018pix2code}.
However, image captioning only predicts what elements are in a UI, but not the bounding box of these elements.
Therefore, pix2code assumes UIs use a small set of predefined fonts, shapes, sizes and layouts, to compensate for this shortage.
However, due to this limitation, our experiments showed that pix2code does not work on real-app UIs which use much more diverse fonts/shapes/sizes/layouts.
In comparison, in this work, we study the object detection methods, which directly obtain the bounding box of the elements.
We further propose a novel top-down coarse-to-fine method to detect elements.

There are also some works~\cite{nguyen2015reverse, beltramelli2018pix2code} which perform object detection on other Platforms (e.g. iOS, website), and other kinds of design (e.g. sketch).
In this work, we perform our experiments on Android app UIs because of the availability of large-scale Rico dataset, while other datasets like REMAUI
cannot effectively train and experiment deep learning models (see our results of training data size in Section~\ref{sec:trainingdatasize}).
However, since our model does not make any specific assumptions about Android UIs, we believe that our model could be easily generalized to other platforms and other kinds of GUI design.
We leave them as future work because they demand significant manual labelling effort.
We release \href{http://uied.online/}{\textcolor{blue}{our tool}} to public.

\section{Conclusion}
This paper investigates the problem of GUI element detection.
We identify four unique characteristics of GUIs and GUI elements, including large in-class variance, high cross-class similarity, packed or close-by elements and mix of heterogeneous objects.
These characteristics make it a challenging task for existing methods (no matter old fashioned or deep learning) to accurately detect GUI elements in GUI images.
Our empirical study reveals the underperformance of existing methods borrowed from computer vision domain and the underlying reasons, and identifies the effective designs of GUI element detection methods.
Informed by our study findings, we design a new GUI element detection approach with both the effective designs of existing methods and the GUI characteristics in mind.
Our new method achieves the state-of-the-art performance on the largest-ever evaluation of GUI element detection methods.

\begin{acks}
This research was partially supported by the Australian National University Data61 Collaborative
Research Project(CO19314) and Facebook gift funding.
\end{acks}

\bibliographystyle{ACM-Reference-Format}
\bibliography{reference}

\end{document}